\documentclass[11pt,a4paper]{article}
\usepackage[utf8]{inputenc}
\usepackage[T1]{fontenc}
\usepackage{geometry}
\usepackage{graphicx}
\usepackage{listings}
\usepackage{xcolor}
\usepackage{chngcntr}
\usepackage{authblk} 
\usepackage{hyperref} 

\hypersetup{
    colorlinks=true,
    linkcolor=blue,
    citecolor=blue,
    urlcolor=blue
}

\AtBeginDocument{%
}

\title{Accelerating AI-Powered Research: The PuppyChatter Framework for Usable and Flexible Tooling}

\author[1]{Chun-Hsiung Tseng\thanks{Corresponding author: lendle.tseng.archive@gmail.com}}
\author[2]{Hao-Chiang Koong Lin}
\author[3]{Andrew Chih-Wei Huang}
\author[4]{Yung-Hui Chen}
\author[1]{Jia-Rou Lin}

\affil[1]{Dept. of Electrical Engineering, YuanZe Univ., Taoyuan, R.O.C.}
\affil[2]{Dept. of Information and Learning Technology, National Univ. of Tainan, Tainan, R.O.C.}
\affil[3]{Dept. of Psychology, Fo Guang Univ., Yilan, R.O.C.}
\affil[4]{Dept. of Computer Information and Network Engineering, Lunghwa Univ., Taoyuan, R.O.C.}

\date{} 

\begin{document}

\maketitle

\begin{center}
\small
\textit{This is the author's version of the work. The definitive version was presented at the 14th International Conference on Ubi-Media Computing (Umedia 2026) and is scheduled to be published by Springer in the Communications in Computer and Information Science (CCIS) series volume.}
\end{center}
\vspace{1.5em}

\begin{abstract}
This research addresses the challenges inherent in developing Artificial Intelligence (AI) applications, particularly those leveraging Large Language Models (LLMs). While AI vendors provide Application Programming Interfaces (APIs) and Software Development Kits (SDKs) to facilitate developer interaction, the former often requires intricate manual request construction, and the latter can lead to significant vendor lock-in. Furthermore, existing model abstraction frameworks, though mitigating vendor dependency, introduce an additional layer of complexity and potential security concerns. To reconcile these conflicting factors, the study introduces PuppyChatter, a novel software framework designed to preserve the intuitive simplicity of vendor-specific SDKs while simultaneously adhering to the vendor-neutrality principles characteristic of model abstraction, thereby offering a more streamlined and flexible development paradigm.

\vspace{0.5em}
\end{abstract}

\section{Motivation}
The emergence of Artificial Intelligence (AI) has garnered significant attention across numerous fields, including academic research, software development, and social interactions. AI is now widely applied to a variety of tasks, such as searching and summarizing information, aiding business decisions, performing translations, generating images, and planning daily activities. While end-users often interact with Large Language Models (LLMs) through simple, prompt-based interfaces, developers are afforded significantly more flexibility via Application Programming Interfaces (APIs). These APIs are essential tools provided by AI vendors for harnessing the full potential of LLMs within custom research and software applications. AI service providers that offer Application Programming Interfaces (APIs) require requests to conform to their specified formats. For instance, to access the ChatGPT series of models, a developer must construct a JSON payload that adheres to the OpenAI API specification and transmit it to the designated endpoint via the HTTP protocol.

Manually constructing raw requests, as described, is an error-prone and time-consuming task. To mitigate these issues, most AI service providers offer dedicated software libraries to simplify this process. Using these libraries enables developers to write more comprehensible and maintainable code. Furthermore, these Software Development Kits (SDKs) facilitate the construction of practical applications. Conversely, relying on vendor-specific SDKs can result in vendor lock-in, potentially restricting architectural flexibility and increasing long-term dependency. To alleviate this issue, model abstraction frameworks, such as LangChain\footnote{https://www.langchain.com/}, are often adopted. Such frameworks establish an abstract layer over various LLM SDKs, thereby decoupling applications from their specific underlying implementations. However, these frameworks introduce an additional layer of complexity. As Mavroudis \cite{Mavroudis2025} highlights, while their emphasis on modularity and integration is a strength for development, these frameworks also "introduce complexities and potential security concerns that warrant critical examination."

This research endeavors to harness the respective strengths of two distinct approaches while simultaneously addressing and alleviating their inherent limitations. More specifically, it seeks to integrate the user-friendly simplicity characteristic of the native-vendor-SDK approach with the broader vendor-neutrality inherent in the model abstraction paradigm. To this end, we introduce PuppyChatter, a novel software framework meticulously designed to adhere to the robust principles of the model abstraction paradigm, thereby effectively preserving the intuitive simplicity often associated with the vendor SDK approach. The subsequent sections of this manuscript are structured to provide a comprehensive overview, commencing with a summary of pertinent related literature, followed by a detailed exposition of PuppyChatter's design, illustrative examples of its practical application, and concluding with a discussion of implications and prospective future directions.

\section{Related Works}
With the emergence of Artificial Intelligence (AI), many AI SDKs have been developed and adopted in diverse fields. A prominent example is the \textit{OpenAI SDK}. Its ChatGPT is widely used for tasks such as text generation, summarization, and translation\cite{Alhur2024}. OpenAI Codex, the model behind GitHub Copilot, translates natural language into code, supporting over 10 programming languages. It is used for building websites, translating programming languages, and answering data science queries\cite{KumarSharma2023}. \textit{Gemini SDK} is another example. Gemini Advanced is utilized in healthcare for data analytics, early disease detection, and supporting medical decision-making. It plays a significant role in enhancing patient care and treatment protocols\cite{Alhur2024}. In the field of software development, a study comparing code generation capabilities for Human-Robot Interaction, existing AI SDKs achieved a 60\% success rate, highlighting their potential in specialized programming tasks\cite{Sobo2025}. While these AI SDKs offer significant advancements, the integration of AI technologies requires careful consideration of ethical standards and the maintenance of system reliability and accuracy\cite{McIntosh2025}.

\section{System Design}
\begin{figure}
    \centering
    \includegraphics[width=0.75\linewidth]{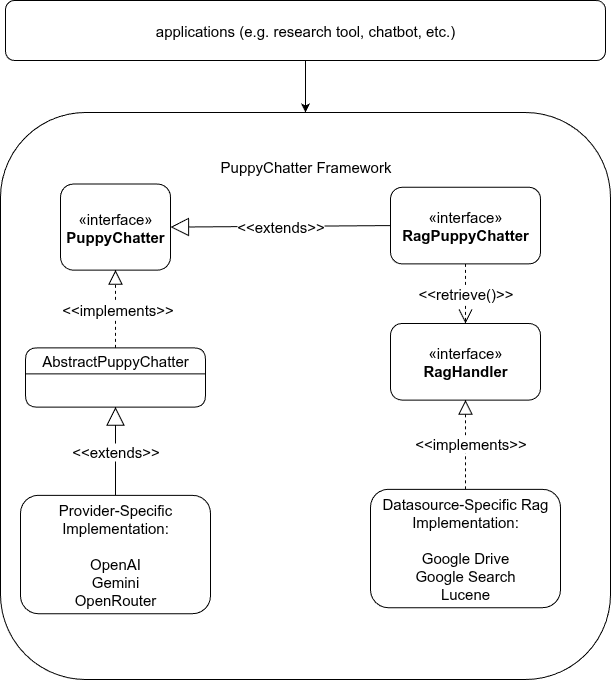}
    \caption{System Architecture}
    \label{fig:systemArchitecture}
\end{figure}

Addressing the inherent challenges of usability and flexibility in developing AI-powered applications, we developed \textit{PuppyChatter}. This lightweight Java framework is specifically designed with an emphasis on simplicity and approachability, qualities reflected in its name. The design philosophy is centered on three core principles: 
\begin{enumerate}
    \item Clear Separation of Concerns
    \item High Extensibility through Abstraction
    \item Pragmatic Integration of Information Retrieval Techniques.
\end{enumerate}
Figure \ref{fig:systemArchitecture} illustrates the system architecture of \textit{PuppyChatter}.

\subsection{Core Interaction Layer: The PuppyChatter Interface}
\textit{PuppyChatter} is the main interface of the framework. Listing \ref{lst:puppychatter} shows some important methods declared in the interface.
\begin{lstlisting}[caption={PuppyChatter Interface Methods}, label={lst:puppychatter}] 
public interface PuppyChatter<T extends PromptParameters, S extends Response> {
public S bark(String sessionId, String prompt) throws BarkException;
public void bark(String sessionId, String prompt, BarkCallback barkCallback) throws BarkException;
public S bark(String sessionId, String prompt, T parameters) throws BarkException;
public void bark(String sessionId, String prompt, T parameters, BarkCallback barkCallback) throws BarkException;
}
\end{lstlisting}
Note that methods with a \textit{BarkCallback} object as an argument will run in an asynchronous manner.

\subsection{Pluggable Model Providers}
A key design goal of \textit{PuppyChatter} is to mitigate vendor lock-in and provide developers with the flexibility to switch between different Large Language Models (LLMs) based on cost, performance, or specific features. This is achieved by offering concrete implementations of the \textit{AbstractPuppyChatter} class for a variety of popular LLM providers and endpoints. This modular, "pluggable" architecture ensures that an application can adapt to the rapidly changing AI landscape with minimal code alterations, often only requiring the instantiation of a different \textit{PuppyChatter} implementation. The framework currently includes built-in support for the following providers: 
\begin{enumerate}
    \item OpenAI-compatible APIs: The \textit{OpenAICompatiblePuppyChatter} class provides a generic connector for any endpoint that adheres to the OpenAI API specification.
    \item Google Gemini: \textit{GeminiAqaPuppyChatter} is a specialized implementation tailored for Google's Gemini family of models. 
    \item Model Routers: To grant access to an extensive array of models through a unified interface, the \textit{OpenrouterPuppyChatter} class is provided. 
    \item Local Models: For usage scenarios requiring offline operation, data privacy, or fine-tuned local models, \textit{PuppyChatter} includes \textit{OllamaPuppyChatter}.
\end{enumerate}

\subsection{Extensible Retrieval-Augmented Generation (RAG)}
Retrieval-Augmented Generation (RAG) has emerged as a crucial paradigm in natural language processing, significantly enhancing the capabilities and reliability of large language models (LLMs). While vector stores are frequently championed as a cornerstone for RAG systems, their implementation often introduces significant complexity that may not always be warranted. According to Park et al., the complexity of vector stores can be a barrier to efficient RAG implementation, especially in resource-constrained environments like mobile or edge devices\cite{Park2025}. PuppyChatter provides a highly extensible and pragmatic approach to RAG, primarily through its \textit{RagHandler} interface and a Decorator pattern implemented by \textit{RagPuppyChatter}. This design cleanly separates the retrieval logic from the LLM's generative process. Listing \ref{lst:raghandler} shows the definition of \textit{RagHandler}.
\begin{lstlisting}[caption={RagHandler Interface Methods}, label={lst:raghandler}] 
public interface RagHandler {
    public List<Chunk> getChunks(String sessionId, PromptParameters parameters, List<Conversation> messages) throws Exception;
    public Conversation transformLastPrompt(String sessionId, PromptParameters parameters, List<Conversation> messages, List<Chunk> chunks);
}
\end{lstlisting}
A key distinction is its first-class support for traditional keyword-based information retrieval, specifically leveraging Apache Lucene. This offers a robust, interpretable, and lower-barrier alternative or complement to vector database-driven semantic search, making RAG more accessible for researchers with existing keyword-indexed knowledge bases. The framework also includes various data-source-specific \textit{RagHandler} implementations for common sources like Google Drive, Google Search, and Tavily, allowing for diverse and hybrid retrieval strategies. Note that to prevent the potentially inaccurate issue of keyword-indexed knowledge databases, one can still extend from the \textit{RagHandler} interface to use vector databases or use an AI-backed knowledge base such as Tavily.

\subsection{Caching and Extensibility}
To improve performance and reduce operational costs, the framework includes a caching mechanism abstracted by the \textit{CacheService} interface. The default implementation, \textit{FileSystemCacheService}, provides a simple file-based cache for storing conversation histories and model responses.
Finally, for asynchronous operations and custom logic, the \textit{BarkCallback} interface allows developers to implement non-blocking handlers for model responses, enabling seamless integration into reactive application architectures.

\section{Usage Scenarios}
\begin{figure}
    \centering
    \includegraphics[width=0.75\linewidth]{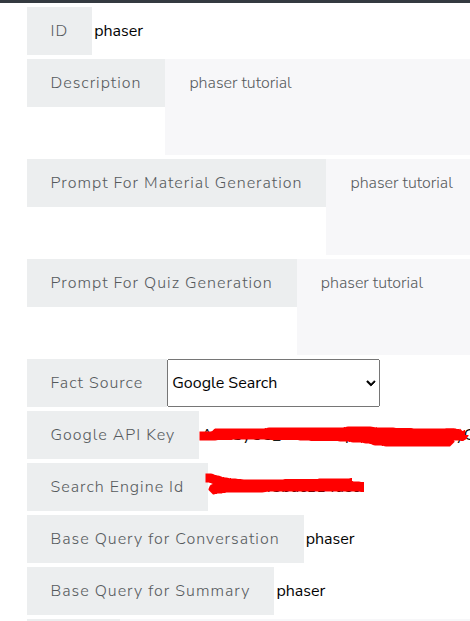}
    \caption{Knowledge Source Configuration of PuppyChatterWeb}
    \label{fig:puppychatterweb_ds}
\end{figure}

In this section, we showcase the usage scenario of \textit{PuppyChatter} by introducing two extended works: \textit{PuppyChatterWeb} and \textit{PuppyCodeReview}, both developed upon its foundation. Specifically, \textit{PuppyChatterWeb} functions as a framework designed for the construction of pedagogical AI agents \cite{Lin2025}. Leveraging the RAG functionalities of \textit{PuppyChatter}, this framework offers a streamlined method for end-users to configure their knowledge sources. Figure \ref{fig:puppychatterweb_ds} illustrates an example utilizing the Google Search implementation of the \textit{RagHandler}. Subsequently, \textit{PuppyChatterWeb} is capable of generating learning outlines and sequential steps for a specified topic, drawing upon the configured knowledge source.

On the other hand, \textit{PuppyCodeReview} is an AI-based code review system\cite{Tseng2025}. The system's code review process integrates predefined software engineering rules with the \textit{PuppyChatter} library. These rules, detailed in the \textit{prompt.txt} file, comprise a comprehensive set of guidelines ranging from high-level design principles to specific "code smells" such as "Long Method" or "Feature Envy." The \textit{DefaultPuppyCodeReviewer} class facilitates this process by combining the student's code with the \textit{prompt.txt} rules before transmitting this integrated package to an \textit{OpenrouterPuppyChatter} instance. The \textit{PuppyChatter} library then utilizes these detailed instructions to perform a comprehensive code analysis, moving beyond mere correctness checks to evaluate against established software engineering best practices. The system subsequently generates a structured JSON response, providing scores across various metrics and specific feedback, all guided by the initial \textit{prompt.txt} ruleset.

\section{Acknowledgement}
This research is partially supported by the 114-2410-H-155-012-MY2 project, which was funded by the National Science and Technology Council, Taiwan, R.O.C.

\section{Conclusions and Future Work}
This paper introduces \textit{PuppyChatter}, a lightweight Java framework designed to reconcile the inherent conflict in AI application development: balancing the simplicity of vendor-specific SDKs with the flexibility of vendor-neutral abstraction layers. \textit{PuppyChatter} achieves this by offering an intuitive, SDK-like developer experience while ensuring architectural adaptability through its pluggable design. The framework's core contributions lie in its distinct separation of concerns, extensive extensibility via abstraction, and a pragmatic approach to RAG. By supporting diverse LLM providers and facilitating accessible RAG implementations, including traditional keyword-based retrieval, \textit{PuppyChatter} significantly lowers the barrier to developing sophisticated AI tools. Demonstrated through usage scenarios such as \textit{PuppyChatterWeb} and \textit{PuppyCodeReview}, the framework establishes a robust foundation for practical, real-world applications in education and software engineering. Future work will expand the library of supported LLM providers and \textit{RagHandler} implementations to enhance its versatility. Additionally, we will explore advanced caching strategies and integrate agentic workflows to support more complex, automated tasks, thereby solidifying \textit{PuppyChatter}'s role as a key enabler for flexible and efficient AI development.

\end{document}